\PassOptionsToPackage{table}{xcolor}
\documentclass[11pt]{article}

\usepackage[final]{acl}

\usepackage{times}
\usepackage{latexsym}

\usepackage[T1]{fontenc}

\usepackage[utf8]{inputenc}

\usepackage{microtype}

\usepackage{inconsolata}

\usepackage{graphicx}
\usepackage{amsfonts}
\usepackage{booktabs}
\usepackage{multirow}
\usepackage{xcolor}
\usepackage{arydshln}
\usepackage{tabularx}
\usepackage{array}
\usepackage{tabularx}
\usepackage{makecell}
\usepackage{amsmath}
\usepackage{amssymb}
\usepackage{float}
\usepackage{tikz}
\usetikzlibrary{shadows}
\usepackage[ruled,vlined]{algorithm2e}
\usepackage{hyperref}
\usepackage{fontawesome5}

\newcolumntype{M}[1]{>{\raggedright\arraybackslash\fontsize{7.6pt}{9.2pt}\selectfont}m{#1}}
\newcolumntype{A}[1]{>{\centering\arraybackslash}m{#1}}
\newcolumntype{Y}{>{\centering\arraybackslash}X}
\newcolumntype{L}[1]{>{\raggedright\arraybackslash}m{#1}}

\newcommand{\tightmidrule}{\specialrule{\lightrulewidth}{0.45ex}{0pt}}
\DeclareMathSizes{7.2}{7.2}{5}{5}

\definecolor{TableGray}{RGB}{240,240,240}
\definecolor{OursBlue}{RGB}{201,218,243}
\definecolor{GroupBlue}{RGB}{232,242,252}

\title{
\makebox[0pt][r]{%
  \raisebox{-0.35\height}{\includegraphics[height=1.2cm]{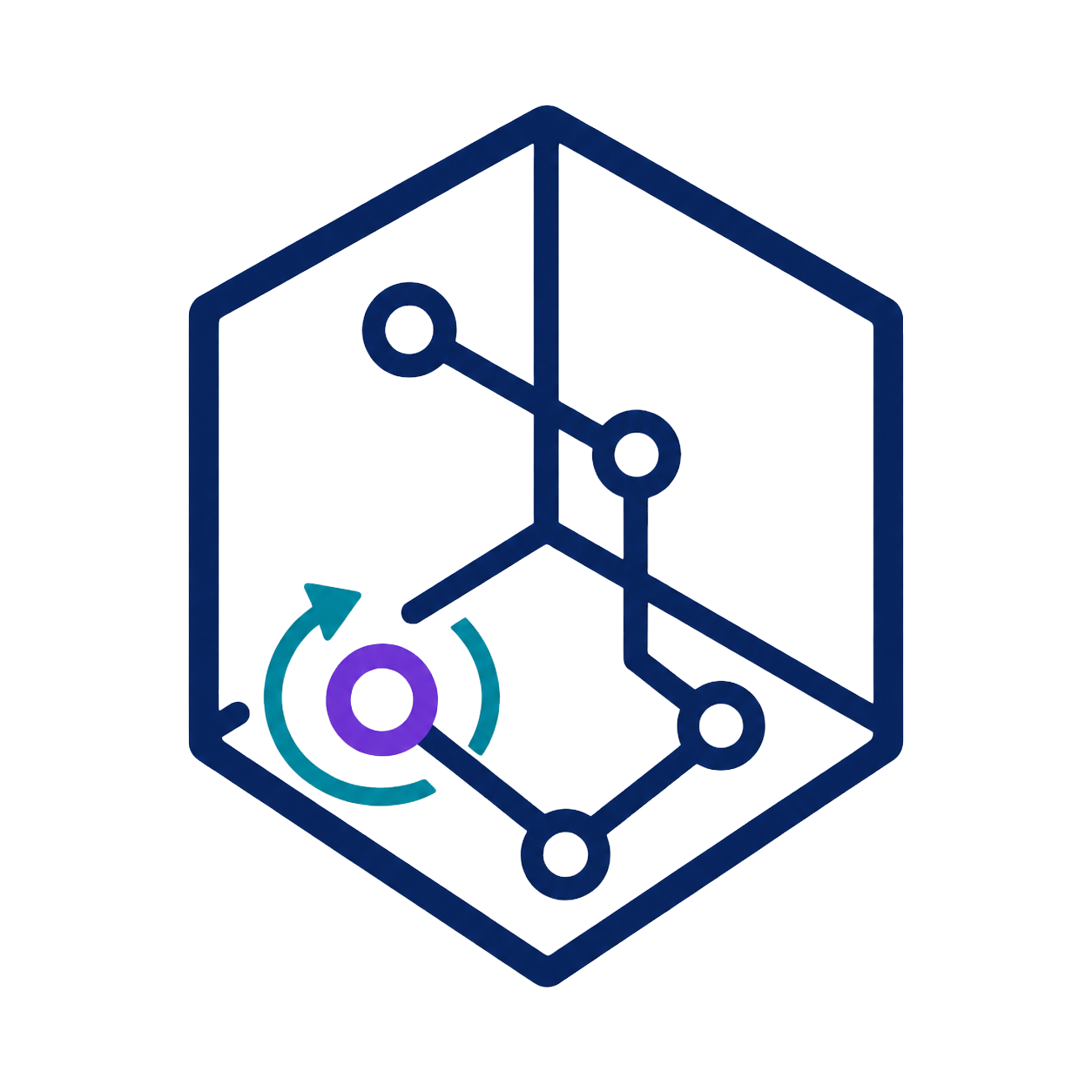}}%
  \hspace{0em}%
}%
LongSpace: Exploring Long-Horizon Spatial Memory \\
from Perception to Recall in Video
}

\author{
 \textbf{Shiqiang Lang\textsuperscript{1,2}},
 \textbf{Jing Liu\textsuperscript{2,3}},
 \textbf{Haoyang He\textsuperscript{1}},
 \textbf{Peiwen Sun\textsuperscript{4}},
 \textbf{Yuanteng Chen\textsuperscript{2,3}},
\\
 \textbf{Tao Liu\textsuperscript{2,5}},
 \textbf{Lan Yang\textsuperscript{1}\thanks{Corresponding authors.}},
 \textbf{Longteng Guo\textsuperscript{2,3}\footnotemark[1]},
 \textbf{Honggang Zhang\textsuperscript{1}}
\\
 \textsuperscript{1}Beijing University of Posts and Telecommunications,
 \textsuperscript{2}Zhongguancun Academy,
 \\
 \textsuperscript{3}Institute of Automation, Chinese Academy of Sciences,
 \\
 \textsuperscript{4}The Chinese University of Hong Kong,
 \textsuperscript{5}Xi'an Jiaotong University
 \\
\normalsize{
\faGithub\ \href{https://github.com/ShiqiangLang/LongSpace}{\texttt{GitHub}}
}
}

\definecolor{geminiBlue}{HTML}{8E8ED7}
\definecolor{qwenBlue}{HTML}{78A2E0}

\definecolor{myred}{rgb}{0.7, 0.3, 0.0}
\definecolor{myblue}{HTML}{0a41b8}
\definecolor{mygreen}{HTML}{056b34}
\definecolor{mypurple}{HTML}{5d1e8b}
\definecolor{TableGray}{RGB}{240,240,240}
\definecolor{OursBlue}{RGB}{201,218,243}
\definecolor{lightpurple}{RGB}{242,238,250}

\begin{document}
\maketitle
\begin{abstract}
Multimodal Large Language Models (MLLMs) have advanced image and video understanding and can increasingly handle longer visual inputs. Long-horizon tasks such as autonomous driving and robotic navigation require more than recognizing the current view, as models must remember and retrieve previously observed spatial layouts, routes, viewpoint changes, and object states. To evaluate this capability, we introduce LongSpace-Bench, a room-tour video benchmark for long-horizon spatial memory, covering scene perception, spatial relations, and spatial memory. In this work, we further propose LongSpace, a memory framework for long-video spatial reasoning. LongSpace models long videos as sequential chunks, incorporates 3D structural cues into early decoder layers, and constructs layer-aware memory for question-guided retrieval. Experiments on multiple spatial reasoning benchmarks show that LongSpace improves long-video spatial understanding, further demonstrating explicit spatial memory as a key capability for long-horizon video MLLMs.
\end{abstract}

\begin{figure*}[t]
    \centering
    \includegraphics[width=\textwidth,page=1]{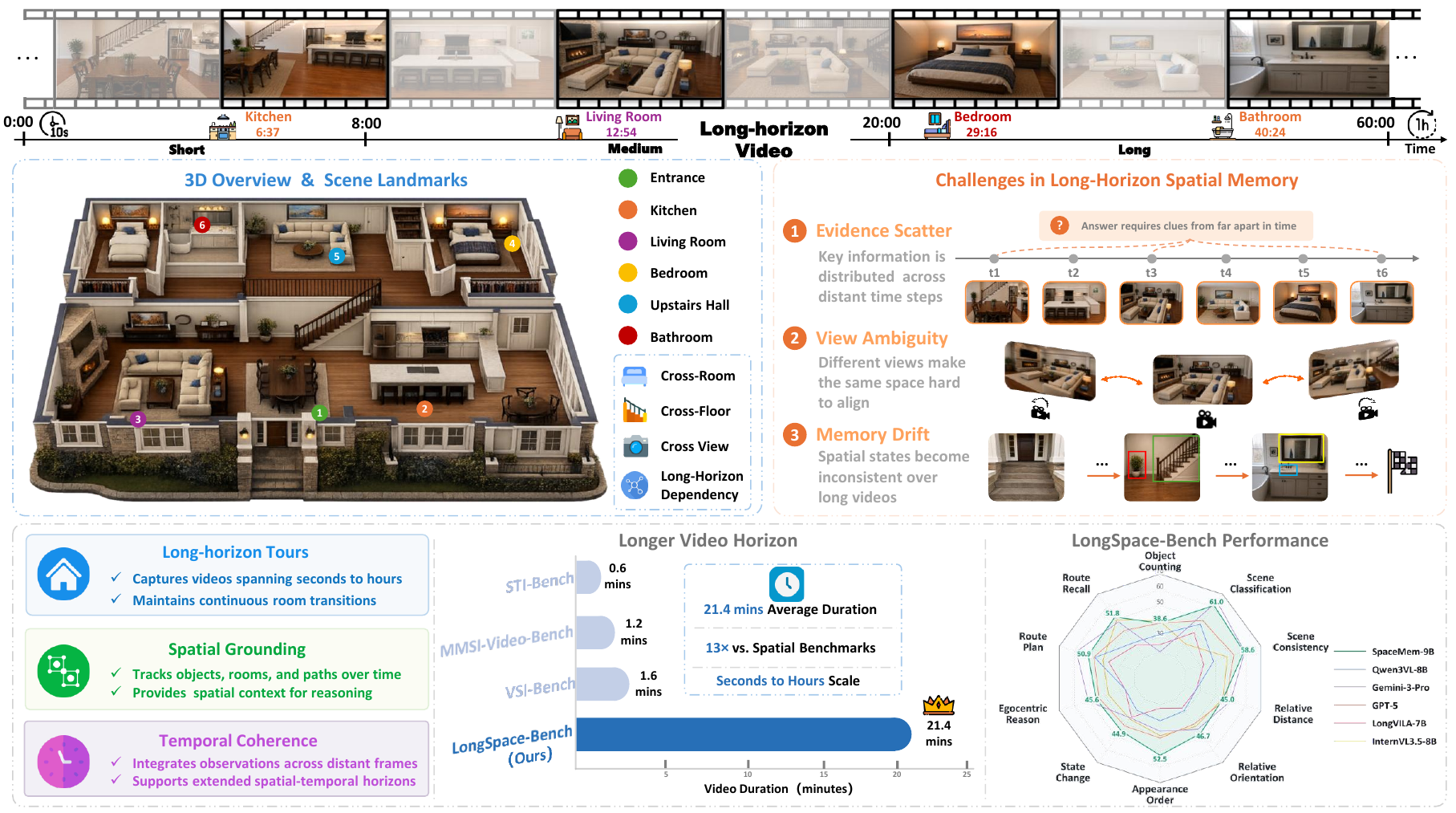}
    \vspace{-2.00em}
    \caption{Long-horizon spatial memory require spatial evidence to be retained across distant observations, changing views, and evolving scene states. LongSpace-Bench spans video horizons from seconds to hours and evaluates spatial perception, relations, and memory over continuous room-tour videos.
    }
    \label{fig:teaser}
    \vspace{-1.15em}
\end{figure*}

\section{Introduction}
\label{sec:introduction}
Recent MLLMs are extending visual understanding from static images to longer visual inputs~\citep{zhang2024long,qian2024streaming,zhang2025videollama,chen2025longvila}. In continuous visual observations, spatial memory is a central capability. Models must not only recognize visible objects and events, but also maintain an understanding of scene layouts, object relationships, viewpoint changes, and navigable structures over time. In applications such as autonomous driving, robotic navigation, and embodied assistance, later decisions or questions often depend on spatial evidence observed much earlier. Recent studies have evaluated or improved spatial reasoning in multi-image, multi-view, and video settings~\citep{yang2025mmsibench,xu2025multispatialmllm,yeh2025allangles,lin2025mmsivideobench}, but most work still focuses on short-term contexts or local spatial relations.

As the observation horizon extends, spatial reasoning increasingly relies on long-horizon spatial memory, as illustrated in Figure~\ref{fig:teaser}. Unlike general temporal understanding, spatial evidence exhibits structural persistence, with layouts, depth cues, orientations, and route relations distributed across different segments that can influence later answers even when absent from the current visual context. Simply increasing the input length is insufficient, as redundant visual tokens may dilute important spatial cues, and unstructured segment-level information throughout the video is difficult to retrieve for subsequent reasoning. Effective long-horizon spatial reasoning thus benefits from models capable of extracting reliable spatial cues from local observations while maintaining these cues over extended temporal intervals for question-guided retrieval. In practice, long-horizon spatial memory represents the capacity to retain, organize, retrieve, and utilize spatial evidence across prolonged observations.

However, existing evaluations remain limited in capturing this capability. They often focus on short videos, multi-image inputs, or local relations~\citep{li2025stibench,zhu2026videomsr}, and seldom address both \textbf{long-horizon} observations and \textbf{multi-dimensional} spatial abilities simultaneously. To address this gap, we introduce LongSpace-Bench, a room-tour video benchmark for long-horizon spatial memory. Constructed from real-world room-tour videos, LongSpace-Bench encompasses continuous indoor layouts, room transitions, object arrangements, and navigation routes, aligning with the requirement to preserve and retrieve long-range spatial evidence outlined above. Its tasks span three levels: scene perception, spatial relations, and spatial memory, including recognition of stable scene semantics, assessment of geometric relations such as distance and orientation, and memory-intensive reasoning over appearance order, state changes, route planning, and route recall. Collectively, these tasks enable LongSpace-Bench to evaluate whether models can retain, organize, retrieve, and utilize spatial information over extended temporal horizons.

To enable long-horizon spatial reasoning and memory, we further propose LongSpace. It directly addresses the modeling requirements outlined above by obtaining reliable spatial cues from local observations and preserving them for retrieval across segments. Prior studies indicate that geometry-enhanced models facilitate the capture of depth, orientation, and layout~\citep{fan2025vlm,zheng2026learning,zhao2025spacemind}, while research on spatial memory highlights the importance of storing and reusing long-term scene information in a structured form~\citep{yang20253d,cai2025vision,hu20263dllm}. Building on these insights, LongSpace integrates geometry-aware perception with retrievable memory. It represents a long video as an ordered sequence of chunks, aligns geometry features within each chunk to strengthen local spatial representations, and constructs retrievable layer-aware memory that maintains structured spatial evidence across chunks. During question answering, the model retrieves relevant evidence from memory to guide generation. LongSpace thus aims not only to increase the number of input frames, but to establish a queryable long-horizon spatial memory throughout the observation sequence. Experimental results demonstrate that LongSpace produces larger improvements on memory-intensive tasks, indicating that explicit spatial memory delivers advantages beyond stronger visual encoders or extended input contexts.

Our contributions are summarized as follows:
\begin{itemize}
    \item We introduce \textbf{LongSpace-Bench}, a benchmark for evaluating long-video spatial reasoning and memory over real-world room-tour videos.
    \item We propose \textbf{LongSpace}, a framework that integrates geometry-aware perception with retrievable long-horizon video memory.
    \item We conduct a comprehensive evaluation on LongSpace-Bench across proprietary, open-source, and spatial-centric models, providing empirical analysis of long-video spatial memory.
\end{itemize}

\begin{table*}[!t]
    \centering
    \scriptsize 
    \setlength{\tabcolsep}{3.6pt}
    \renewcommand{\arraystretch}{1.08}

    \newcommand{\lvyes}{\includegraphics[height=0.9em]{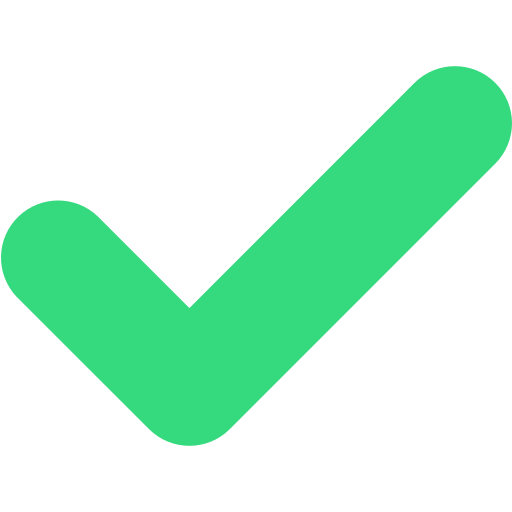}}
    \newcommand{\lvno}{\includegraphics[height=1.0em]{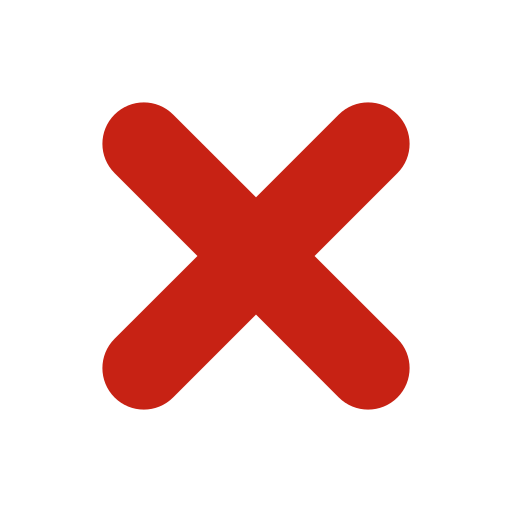}}
    
    \caption{Comparison of LongSpace-Bench with representative spatial reasoning and memory benchmarks. MC, NA, and OE denote multiple-choice, numerical-answer, and open-ended formats. Avg. Duration denotes average video duration; -- indicates unavailable or not applicable statistics. Multi-Dim. Eval. indicates whether a benchmark covers multiple spatial ability dimensions rather than a narrow relation type.}
    \vspace{-1.25em}
    \label{tab:benchmark_comparison}
    \resizebox{\textwidth}{!}{%
    \begin{tabular}{@{}lccccccccc@{}}
        \toprule
        \multicolumn{1}{c}{\multirow[c]{2}{*}{\textbf{Benchmark}}} &
        \multirow[c]{2}{*}{\textbf{Modality}} &
        \multicolumn{3}{c}{\textbf{Scale}} &
        \multicolumn{3}{c}{\textbf{Benchmark Design}} &
        \multirow[c]{2}{*}{\makecell[c]{\textbf{Long} \\ \textbf{Video}}} &
        \multirow[c]{2}{*}{\makecell[c]{\textbf{Multi-Dim.} \\ \textbf{Eval.}}} \\
        \cmidrule(lr){3-5}\cmidrule(lr){6-8}
        & & \textbf{Samples Num} & \textbf{QA Pairs} & \makecell[c]{\textbf{Avg.} \\ \textbf{Duration}} & \textbf{Annotation} & \textbf{Tasks} & \textbf{Answer} & & \\
        \midrule
        SpatialRGPT~\citep{cheng2024spatialrgpt} & Image & 1,406 & 1,406 & -- & Auto & 2 & OE+NA & \lvno & \lvno \\
        SpatialVLM~\citep{chen2024spatialvlm} & Image & 546 & 546 & -- & Auto\&Human & 2 & OE+NA & \lvno & \lvno \\
        CVBench~\citep{tong2024cambrian} & Image & 2,638 & 2,638 & -- & Auto & 2 & MC & \lvno & \lvno \\
        All-Angles-Bench~\citep{yeh2025allangles} & Image & 380 & 2,132 & -- & Human & 6 & MC & \lvno & \lvno \\
        MMSI-Bench~\citep{yang2025mmsibench} & Image & 1,990 & 1,000 & -- & Human & 11 & MC & \lvno & \lvyes \\
        SPAR-Bench~\citep{zhang2026flatland} & Image & 14,708 & 7,207 & -- & Auto\&Human & 20 & MC+NA/OE & \lvno & \lvyes \\
        VSI-Bench~\citep{yang2024thinking} & Video & 288 & 5,000+ & 1.2mins & Auto\&Human & 8 & MC+NA & \lvno & \lvyes \\
        STI-Bench~\citep{li2025stibench} & Video & 369 & 2,064 & 0.6mins & Auto\&Human & 8 & MC & \lvno & \lvyes \\
        MMSI-Video-Bench~\citep{lin2025mmsivideobench} & Video & 1,278 & 1,106 & 1.6mins & Human & 13 & MC & \lvno & \lvyes \\
        \midrule
        \textbf{LongSpace-Bench (Ours)} & \textbf{Video} & \textbf{445} & \textbf{4,073} & \textbf{21.4 min} & \textbf{Human} & \textbf{10} & \textbf{MC+NA} & \lvyes & \lvyes \\
        \bottomrule
        \vspace{-0.8em}
    \end{tabular}}
\end{table*}

\section{Related Work}
\subsection{Visual Spatial Reasoning}
Visual spatial reasoning studies whether a model can form a stable understanding of 3D scene structure, relative spatial relations, and viewpoint changes from images or videos. Recent work moves beyond appearance-driven inference by introducing stronger spatial inductive bias into video MLLMs. Cambrian-1~\citep{tong2024cambrian} revisits vision encoder and connector design from a vision-centric perspective, proposing spatially aware aggregation to preserve high-resolution details important for spatial reasoning. Cambrian-S~\citep{yang2025cambrian} integrates persistent memory, implicit 3D cognition, and predictive sensing into a unified spatial framework. SpaceVista~\citep{sun2025spacevista} extends visual spatial reasoning to all-scale scenarios from millimeters to kilometers, combining structured spatial knowledge, scale-aware modeling, and progressive training for better cross-scene spatial understanding. Another line of work enhances spatial reasoning with RGB-video geometric priors. VLM-3R~\citep{fan2025vlm} learns implicit 3D tokens through instruction-aligned 3D reconstruction, while VG-LLM introduces an explicit 3D geometry encoder to enrich visual representations with structural cues. SpaceMind~\citep{zhao2025spacemind} improves geometry-language interaction through camera-guided modality fusion. Recent studies focus on making geometry more effective for reasoning. Spatial-R1~\citep{ouyang2025spatial} emphasizes task-specific optimization for video spatial reasoning.

\subsection{Memory-Enhanced Spatial Reasoning}
Memory-enhanced spatial reasoning studies how a model preserves, updates, and retrieves scene-structured history during continuous observation. Unlike general video memory, this line of work emphasizes spatial consistency, structured organization, and direct support for downstream reasoning. 3D-Mem~\citep{yang20253d} constructs incrementally updatable scene memory through memory and frontier snapshots, while MTU3D~\citep{zhu2025move} maintains a dynamic spatial memory bank for grounding, scene representation, and exploration. 3DLLM-Mem~\citep{hu20263dllm} couples working memory with episodic memory for long-horizon embodied reasoning. OnlineSI~\citep{liu2026onlinesi} maintains a finite explicit spatial memory for online 3D understanding from video streams, and HIMM~\citep{li2026himm} disentangles episodic and semantic memory for long-horizon exploration and question answering. 3DSPMR~\citep{cai2025vision} further emphasize structured reuse of long-term spatial knowledge through spatialized memory retrieval or unified 3D memory from relational, visual, and geometric cues.
\section{LongSpace-Bench}
\label{sec:longspace_bench}
Spatial reasoning over extended visual observations requires more than recognizing isolated objects or local relations. A model may need to recall which room appeared earlier, how two areas are connected, whether an object state changed, or the path connecting different viewpoints. Existing spatial benchmarks offer useful evaluations for static scenes, multi-image reasoning, and short-video understanding, but they do not directly assess whether video MLLMs can preserve and utilize spatial evidence across long-horizon observations. We introduce LongSpace-Bench to address this limitation.

LongSpace-Bench is built for long-horizon spatial reasoning and memory. It evaluates the use of spatial evidence in continuous video observations and covers complementary abilities, including scene perception, spatial relationships, and spatial memory. As shown in Table~\ref{tab:benchmark_comparison}, existing image and multi-image benchmarks mainly assess static spatial understanding, which offers limited insight into spatial evidence that unfolds over time. Existing video benchmarks introduce dynamic inputs, but they often emphasize short clips, local relations, or narrower task scopes. As a result, they provide limited coverage of whether models can retain, retrieve, and combine spatial evidence over longer temporal spans. LongSpace-Bench targets this evaluation gap by testing the use of different types of spatial information across long-horizon observations.

\subsection{Task Definition}
\label{subsec:motivation_task_definition}
LongSpace-Bench organizes long-video spatial ability into three levels: scene perception, spatial relationship, and spatial memory. Scene perception measures a model's understanding of global environments and stable scene semantics, including Object Counting, Scene Classification, and Scene Consistency. Spatial relationship focuses on geometric relations, such as Relative Distance and Relative Orientation, and requires models to infer spatial configurations between objects or regions under viewpoint changes. Spatial memory tests whether models can preserve and retrieve spatial evidence over long temporal horizons, covering Appearance Order, State Change, Egocentric Reasoning, Route Planning, and Route Recall.

\begin{figure}[!t]
    \centering
    \includegraphics[width=\columnwidth,page=1]{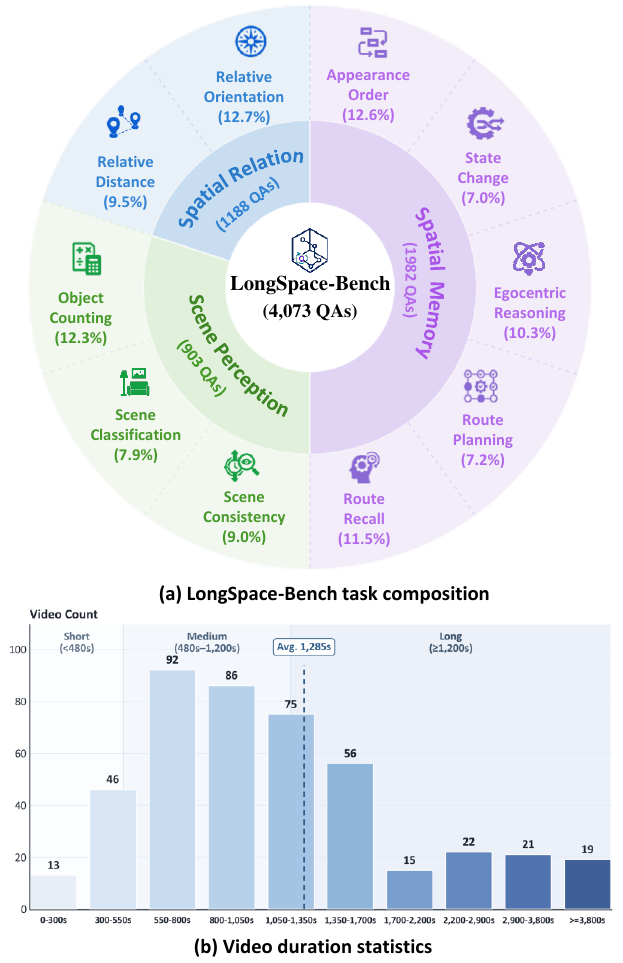}
    \vspace{-2.00em}
    \caption{LongSpace-Bench Statistics Showing (a) Distribution of Question Types Across the Benchmark and (b) Distribution of Video Durations.}
    \label{fig:duration_distribution}
\end{figure}

\subsection{Benchmark Statistic}
\label{subsec:bench_statistic} 
LongSpace-Bench is built from real-world room-tour videos and contains 445 videos, approximately 159 hours of video, and 4,073 question-answer pairs. As shown in Figure~\ref{fig:duration_distribution}, it covers short, medium-, and long-horizon videos and includes ten spatial task types. Object counting uses numerical answers, while the other tasks mainly use multiple-choice answers. Additional annotation details, dataset statistics, benchmark comparison, and evaluation protocol are provided in Appendix Sections~\ref{sec:appendix_bench_construction}--\ref{sec:appendix_experimental_details}.
\section{Method}
\label{sec:method}

\begin{figure*}[t]
    \centering
    \includegraphics[width=\textwidth,page=1]{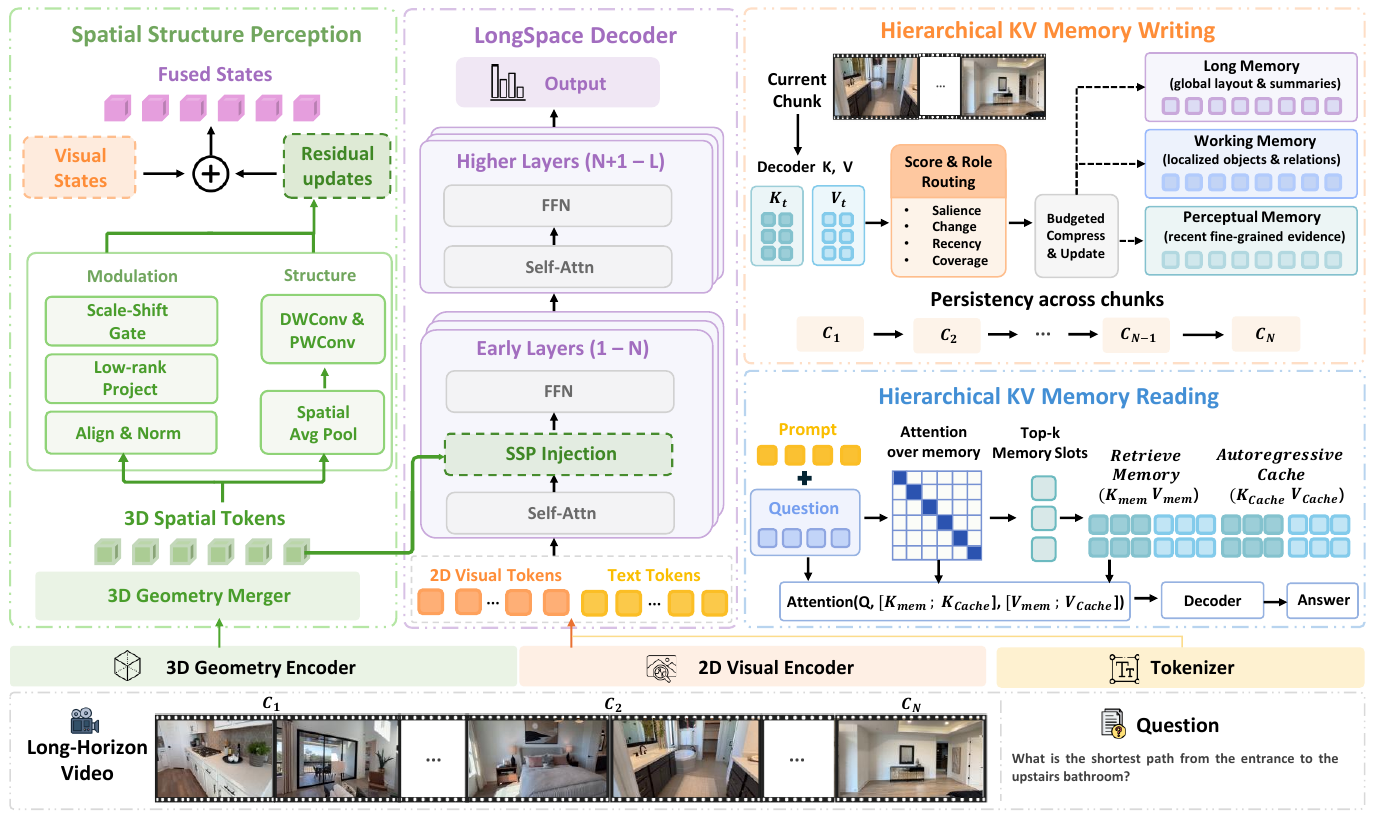}
    \vspace{-2.00em}
    \caption{
    Overview of LongSpace. Spatial Structure Perception fuses 3D spatial tokens with 2D visual representations and injects them into the decoder. Hierarchical KV Memory organizes evidence from sequential video chunks into multi-level memories, which are retrieved according to the question for long-horizon spatial reasoning.
    }
    \vspace{-1em}
    \label{fig:framework}
\end{figure*}

\subsection{Overview}
\label{sec:overview}
Long-horizon spatial reasoning requires models to understand 3D spatial information and retrieve evidence across long temporal ranges. Language-aligned 2D visual features capture semantic appearance but do not explicitly model 3D features. Directly concatenating all video tokens into the language context can exceed the context budget and reduce computational efficiency. LongSpace addresses these issues with spatial geometry-aware perception and layer-wise KV memory. The perception module injects dense 3D features into decoder-side visual tokens, while the memory module compresses selected KV states into layer-wise memories as video segments are encoded temporally. During question answering, LongSpace retrieves query-relevant memories for the final response.

\subsection{Spatial Structure Perception}
\label{sec:ssp}

As shown in Figure~\ref{fig:framework}, SSP fuses Qwen3-VL 2D visual tokens with 3D spatial tokens from the 3D geometry encoder and merger on the decoder side. The 3D spatial features are first pooled, length-aligned, and normalized to match the number of visual tokens $n_v$:
\begin{equation} 
\mathbf{G} = \operatorname{Align} \left(\psi_{\theta}(\Phi_{\mathrm{3d}}(\mathcal{X})), n_v\right). \label{eq:geo_align} \end{equation}

Here, $\mathbf{G}$ denotes the aligned 3D spatial tokens, and $\Phi_{\mathrm{3d}}$ is instantiated with $\pi^3$~\citep{wang2025pi}. SSP maintains $\mathbf{G}$ as an independent spatial-structure stream, rather than merging it only once into the input embeddings.

SSP gathers the current visual states using the visual mask. The visual states and 3D spatial tokens are then normalized and projected into a low-rank bottleneck, which produces a scale term $\gamma$, an offset term $\beta$, and a fusion gate that controls the strength of 3D spatial updates for each visual token. In parallel, SSP extracts a lightweight structural residual by average-pooling the spatial feature map and applying depthwise and pointwise convolutions.

The two paths are then combined as a residual update and written back only to visual-token positions:
\begin{equation}
    \mathbf{H}^{\Omega}_{l}
    \leftarrow
    \mathbf{H}^{\Omega}_{l}
    +
    \Delta_l(\mathbf{H}^{\Omega}_{l}, \mathbf{G}, \mathbf{q}_l).
\label{eq:geo_inject}
\end{equation}

Here, $\mathbf{H}^{\Omega}_{l}$ denotes the hidden states at visual-token positions in layer $l$, $\mathbf{q}_l$ denotes the query context from text states, and $\Delta_l$ denotes the bounded residual from the modulation and structure branches. The updated visual states correspond to the Fused States in the figure and are passed to subsequent layers.

\subsection{Hierarchical KV Memory}
\label{sec:hkm}

Hierarchical KV Memory (HKM) serves as the inference-time memory substrate of LongSpace for preserving long-video evidence under a bounded context. Instead of expanding the final prompt with all video tokens, LongSpace encodes temporally ordered video segments and materializes their evidence as compact hierarchical KV states inside decoder layers. This design enables the final question-answering stage to access cross-segment evidence without keeping the full visual sequence in the language context.

At each decoder layer, HKM turns the current attention states into reusable memory entries, including key-value states, position indices, and hidden features for later selection and compression. These layers capture evidence at different scopes. Sensory layers keep fine-grained visual and spatial evidence, while working layers provide a short-term workspace that binds objects, local spatial relations, and recent changes within the current segment into contextual states. Long-memory layers distill each segment into stable temporal anchors and scene cues, so the final question can locate relevant evidence across long temporal gaps.

For layer $l$ and segment $t$, HKM represents memory update as role-conditioned evidence selection and budget-constrained compression. It first selects candidate evidence from the current segment according to the layer role:
\begin{equation}
    \mathcal{A}_{t,l}
    =
    \operatorname{Select}_{\rho(l)}
    (\mathbf{K}_{t,l},\mathbf{V}_{t,l},\mathbf{F}_{t,l}).
    \label{eq:memory_select}
\end{equation}
Here $\rho(l)$ denotes the layer role, $\mathbf{K}_{t,l}$ and $\mathbf{V}_{t,l}$ are the KV states produced by the current segment, and $\mathbf{F}_{t,l}$ is the hidden feature used for scoring.

It then merges the selected evidence with the previous memory and compresses it under the corresponding role-specific budget:
\begin{equation}
    \mathcal{M}_{t,l}
    =
    \operatorname{Compress}_{\rho(l)}
    (\mathcal{M}_{t-1,l}\cup\mathcal{A}_{t,l};B_{\rho(l)}).
    \label{eq:memory_update}
\end{equation}
Here $B_{\rho(l)}$ is the role-specific memory budget. Since each memory entry retains its position id, the compressed memory preserves the original temporal order of the video.

During memory update, Select treats the KV states, hidden feature, and position id at each video position as a candidate memory entry. It assigns each candidate a priority score using four normalized signals: feature norm for salience, adjacent feature difference for state change, uniformly sampled temporal anchors for coverage, and recency for recent evidence. The signals are combined with role-specific weights from $\rho(l)$. When the number of candidates exceeds the budget, Compress keeps recent or high-scoring entries in raw form and groups the remaining entries in temporal order. Within each group, score-softmax weights are used to pool KV states, features, and positions into compact segment entries. Long-range entries also store segment summaries and segment ids to support retrieval.

\begin{figure}[t]
    \centering
    \includegraphics[width=\columnwidth,height=0.8\columnwidth]{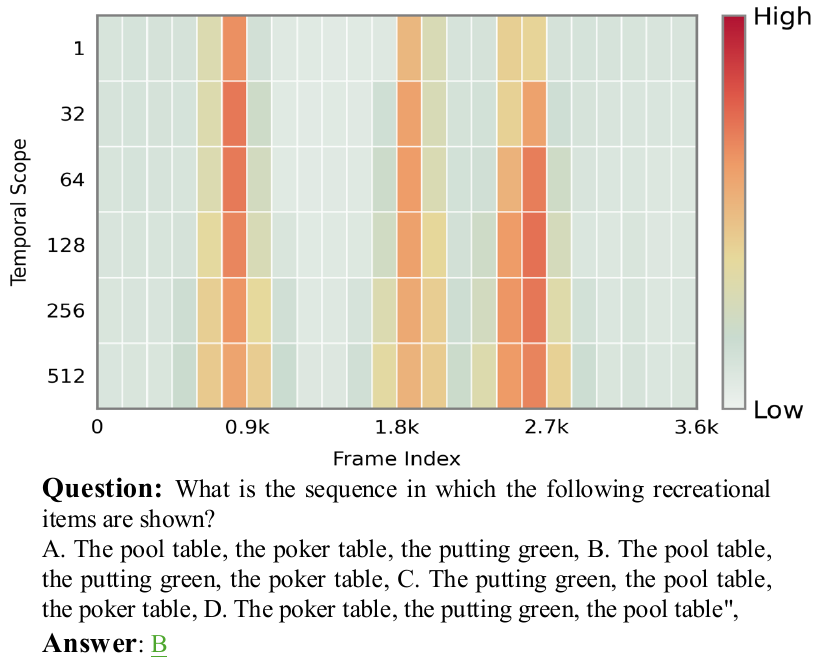}
    \vspace{-1.85em}
    \caption{
    Visualization of LongSpace evidence localization. The heatmap shows how LongSpace identifies sparse question-relevant evidence within hour-level video frames.
    }
    \vspace{-1.9em}
    \label{fig:longspace_visualization}
\end{figure}

\subsection{Memory Retrieval and Decoding}
\label{sec:memory_decoding}

After encoding all video segments, LongSpace encodes the final question as a query and writes it to each HKM layer as the read condition. For sensory and working memories, it performs sparse top-$k$ reading over stored positions. Candidate KV evidence is ranked by both query relevance and memory scores, and the corresponding local KV states are returned. For long memory, LongSpace adopts a segment-to-token coarse-to-fine read: it first matches the query with segment summaries or spatial prototypes to find relevant segments, and then reads compact KV entries from them. This process supplies the decoder with question-relevant hierarchical video evidence without expanding the full video memory. Figure~\ref{fig:longspace_visualization} shows that LongSpace localizes sparse evidence among many irrelevant hour-level video frames.

During decoding, LongSpace keeps the attention operator unchanged and injects the retrieved HKM KV states as a frozen memory prefix. This prefix provides compressed video evidence to each layer, while only the autoregressive cache is updated as new tokens are generated.

\section{Experiments}
\label{sec:experiments}

\begin{table*}[!t]
\centering
\caption{Quantitative comparison of performance on LongSpace-Bench. The best and second-best results among non-proprietary models are highlighted in \textbf{bold} and \underline{underlined}, respectively.}
\vspace{-0.75em}
\label{tab:longspacebench_results}
\setlength{\tabcolsep}{0.6pt}
\renewcommand{\arraystretch}{0.98}
\fontsize{7.2pt}{8.6pt}\selectfont

\newcommand{\subhead}[1]{%
  {\fontsize{5.5pt}{6.9pt}\selectfont\bfseries #1}%
}
\newcommand{\overallhead}{%
  {\fontsize{6.8pt}{8.6pt}\selectfont\bfseries Overall}%
}

\begin{tabular*}{\textwidth}{@{}l@{\extracolsep{\fill}}*{11}{c}@{}}
\toprule
\multicolumn{1}{c}{\multirow[c]{2}{*}{\bfseries Model}} &
\multicolumn{3}{c}{{\bfseries Scene Perception}} &
\multicolumn{2}{c}{{\bfseries Spatial Relation}} &
\multicolumn{5}{c}{{\bfseries Spatial Memory}} &
\multicolumn{1}{c}{\multirow[c]{2}{*}{\overallhead}} \\
\cmidrule(lr{0.28em}){2-4}
\cmidrule(lr{0.28em}){5-6}
\cmidrule(lr{0.28em}){7-11}
&
\subhead{Obj.Cnt.}
& \subhead{Scn.Cls.}
& \subhead{Scn.Cons.}
& \subhead{Rel.Dist.}
& \subhead{Rel.Ori.}
& \subhead{App.Ord.}
& \subhead{St.Chg.}
& \subhead{Ego.Reas.}
& \subhead{Rt.Plan}
& \subhead{Rt.Recall}
& \\
\tightmidrule

\rowcolor{GroupBlue}[0pt][0pt]
\multicolumn{12}{@{}c@{}}{{\bfseries\itshape Proprietary Models}} \\
GPT5~\citep{singh2025openai}
& 38.7 & 49.0 & 52.0 & 46.8 & 39.5 & 41.4 & 41.1 & 43.7 & 47.1 & 41.4 & 43.5 \\
Gemini-3-Pro~\citep{gemini_3_pro_systemcard}
& 20.3 & 64.0 & 59.2 & 49.9 & 43.6 & 37.2 & 34.7 & 50.8 & 58.7 & 48.0 & 45.3 \\

\tightmidrule
\rowcolor{GroupBlue}[0pt][0pt]
\multicolumn{12}{@{}c@{}}{{\bfseries\itshape Open-Source Models}} \\
LongVA-7B~\citep{zhang2024long}
& 14.5 & 46.0 & 42.1 & 40.1 & 30.2 & 28.8 & 35.4 & 29.9 & 43.0 & 31.0 & 32.7 \\
LongVILA-7B~\citep{zhang2024long}
& 25.0 & 39.5 & 32.1 & 27.1 & 26.7 & 22.4 & 34.0 & 26.4 & 37.5 & 29.5 & 29.1 \\
LLaVA-OneVision-1.5-8B~\citep{li2024llava}
& 22.3 & 48.8 & 44.2 & 40.1 & 31.6 & 36.8 & 32.3 & 34.4 & 49.5 & 39.5 & 37.0 \\
LLaVA-NeXT-Video-7B~\citep{liu2024llavanext}
& 9.0 & 39.2 & 30.8 & 26.6 & 30.0 & 31.7 & 28.1 & 33.0 & 37.5 & 38.0 & 29.9 \\
LLaVA-NeXT-Video-72B~\citep{liu2024llavanext}
& 8.8 & 37.9 & 36.4 & 33.6 & 29.5 & 28.0 & 40.4 & 30.6 & 41.6 & 32.4 & 30.5 \\
Qwen2.5-VL-7B~\citep{bai2025qwen25vltechnicalreport}
& 26.4 & 54.2 & 48.6 & 38.0 & 31.2 & 33.9 & 38.2 & 28.7 & 48.1 & 33.1 & 36.7 \\
Qwen2.5-VL-72B~\citep{bai2025qwen25vltechnicalreport}
& 33.1 & \underline{59.7} & 57.3 & \underline{44.7} & \textbf{46.7} & \underline{42.8} & \underline{44.2} & \underline{45.4} & 50.2 & \underline{46.7} & 44.2 \\
Qwen3-VL-8B~\citep{bai2025qwen3}
& 31.5 & 46.3 & 38.6 & 27.1 & 28.9 & 30.5 & 28.4 & 26.4 & 46.4 & 31.6 & 32.9 \\
Qwen3-VL-32B~\citep{bai2025qwen3}
& 36.5 & 55.0 & 54.5 & 43.4 & 37.8 & 41.2 & 42.1 & 39.0 & 48.5 & 44.8 & \underline{46.5} \\
InternVL3.5-8B~\citep{wang2025internvl3}
& 30.5 & 47.7 & 44.8 & 41.4 & \underline{39.9} & 35.6 & 37.4 & 41.1 & 49.8 & 38.4 & 40.0 \\
InternVL3.5-38B~\citep{wang2025internvl3}
& 31.8 & 56.9 & \textbf{58.9} & 43.7 & 39.0 & 38.9 & 42.8 & 39.2 & \textbf{57.7} & 46.5 & 44.2 \\

\tightmidrule
\rowcolor{GroupBlue}[0pt][0pt]
\multicolumn{12}{@{}c@{}}{{\bfseries\itshape Spatial-Centric Models}} \\
Spatial-MLLM-4B~\citep{wu2026spatial}
& 29.9 & 36.5 & 34.0 & 34.6 & 31.2 & 25.5 & 30.2 & 32.1 & 37.9 & 27.3 & 31.4 \\
VG-LLM-4B~\citep{zheng2026learning}
& 23.8 & 45.0 & 37.7 & 37.7 & 33.7 & 31.7 & 38.2 & 34.2 & 47.8 & 39.2 & 36.0 \\
VG-LLM-8B~\citep{zheng2026learning}
& 37.3 & 55.3 & 43.6 & 38.0 & 38.2 & 29.6 & 42.5 & 37.1 & \underline{54.9} & 38.6 & 39.2 \\
VST-3B-SFT~\citep{yang2025visual}
& 16.7 & 45.2 & 34.9 & 36.2 & 38.6 & 33.9 & 31.6 & 39.7 & 44.4 & 33.9 & 34.9 \\
VST-7B-SFT~\citep{yang2025visual}
& 16.1 & 48.0 & 42.1 & 36.7 & 36.8 & 29.8 & 38.9 & 32.8 & 47.1 & 34.1 & 35.0 \\
VLM-3R-7B~\citep{fan2025vlm}
& 37.1 & 42.8 & 45.3 & 42.0 & 36.7 & 38.6 & 42.7 & 40.9 & 41.6 & 39.1 & 40.2 \\
SpatialLadder-3B~\citep{li2025spatialladder}
& 24.4 & 43.1 & 35.2 & 36.4 & 39.9 & 27.8 & 37.2 & 41.8 & 49.5 & 33.7 & 36.0 \\
Cambrian-S-3B~\citep{yang2025cambrian}
& \textbf{47.2} & 44.4 & 45.8 & 37.7 & 33.9 & 32.5 & 35.8 & 30.4 & 44.4 & 32.2 & 37.9 \\
Cambrian-S-7B~\citep{yang2025cambrian}
& \underline{43.4} & 48.5 & 43.9 & 40.8 & 39.7 & 31.7 & 42.1 & 41.6 & 47.1 & 33.0 & 40.5 \\

\tightmidrule
LongSpace-9B (Ours)
& 38.6 & \textbf{61.0} & \underline{58.6} & \textbf{45.0} & \textbf{46.7} & \textbf{52.5} & \textbf{44.9} & \textbf{45.6} & 50.9 & \textbf{51.8} & \textbf{49.2} \\

\bottomrule
\end{tabular*}

\end{table*}

\subsection{Implementation Details}
\label{sec:implementation_details}
\noindent\textbf{Setting.} LongSpace is built on Qwen3-VL-8B~\citep{bai2025qwen3} with $\pi^3$~\citep{wang2025pi} as the 3D geometry encoder, and the geometry-aware module is inserted into the first eight decoder layers. We jointly optimize the language backbone, geometry module, multimodal projector, and language modeling head for one epoch with a global batch size of 64. We use AdamW with a learning rate of $1\times10^{-5}$, a cosine schedule, and a warmup ratio of 0.03. Each video contains at most 32 frames, and all experiments are conducted on 8 NVIDIA A100 80G GPUs.
\\
\noindent\textbf{Training Datasets.} The training data consist of VSI-590K~\citep{yang2025cambrian}, the instruction data introduced by VLM-3R~\citep{fan2025vlm}, and a sampled subset of SPAR-7M~\citep{zhang2026flatland}. These samples cover object-level perception, distance and direction estimation, route reasoning, and appearance-order reasoning.

\subsection{Main Results}
\noindent\textbf{Standard Spatial Reasoning.}We report the results on VSI-Bench~\citep{yang2024thinking} in Table~\ref{tab:vsibench_results}. LongSpace achieves an average score of 70.8, outperforming InternVL3-78B~\citep{wang2025internvl3}, Qwen3-VL-8B~\citep{bai2025qwen3}, and Cambrian-S-7B by 22.4, 12.9, and 7.9 points, respectively. This supports the benefit of geometry-aware perception for local spatial estimation, while LongSpace-Bench tests whether such evidence can be preserved across long videos. Figure~\ref{fig:longspace_visualization} further shows that LongSpace attends to question-relevant spatial regions across temporally separated frames.

\noindent\textbf{Long-Horizon Spatial Memory.}
\begin{table*}[!t]
\centering
\caption{Main results on the VSI-Bench benchmark. Best and second-best scores are highlighted in \textbf{bold} and \underline{underlined}, respectively. Numerical answers are evaluated by MRA and multiple-choice answers by accuracy.}
\vspace{-0.90em}
\label{tab:vsibench_results}
\fontsize{7pt}{9pt}\selectfont
\setlength{\tabcolsep}{1.0pt}
\renewcommand{\arraystretch}{1.0} 

\begin{tabular*}{\textwidth}{@{}l@{\extracolsep{\fill}}*{9}{c}@{\hspace{6pt}}}

\toprule
\multirow{2}{*}{\textbf{Model}} &
\multicolumn{4}{c}{\textbf{\tiny\mbox{Numerical Answer}}} &
\multicolumn{4}{c}{\textbf{\tiny\mbox{Multiple-choice Answer}}} &
\multirow{2}{*}{\textbf{Avg.}} \\
\cmidrule(lr){2-5} \cmidrule(lr){6-9}
& {\tiny\mbox{Obj.Count}} & {\tiny\mbox{Abs.Dist.}} & {\tiny\mbox{Obj.Size}} & {\tiny\mbox{Room Size}}
& {\tiny\mbox{Rel.Dist.}} & {\tiny\mbox{Rel.Dir.}} & {\tiny\mbox{Route Plan}} & {\tiny\mbox{Appr.Order}} & \\

\tightmidrule
\rowcolor{GroupBlue}
\multicolumn{10}{c}{{\bfseries\itshape Proprietary Models}} \\
GPT-5~\citep{singh2025openai}
& 53.3 & 34.4 & 73.3 & 47.5 & 63.7 & 48.6 & \underline{50.2} & 68.9 & 55.0 \\
Gemini-2.5-Pro~\citep{comanici2025gemini}
& 46.0 & 37.4 & 68.7 & 54.4 & 62.0 & 43.9 & 47.4 & 68.8 & 53.6 \\
Gemini-3-Pro~\citep{gemini_3_pro_systemcard}
& 49.0 & 42.8 & 71.5 & 41.8 & 56.6 & 57.5 & \textbf{61.9} & 60.0 & 56.0 \\

\tightmidrule
\rowcolor{GroupBlue}
\multicolumn{10}{c}{{\bfseries\itshape Open-Source Models}} \\
LongVA-7B~\citep{zhang2024long}
& 38.0 & 16.6 & 38.9 & 22.2 & 33.1 & 43.3 & 25.4 & 15.7 & 29.2 \\
VILA-1.5-8B~\citep{lin2024vila}
& 17.4 & 21.8 & 50.3 & 18.8 & 32.1 & 34.8 & 31.0 & 24.8 & 28.9 \\
VILA-1.5-40B~\citep{lin2024vila}
& 22.4 & 24.8 & 48.7 & 22.7 & 40.5 & 25.7 & 31.5 & 32.9 & 31.2 \\
LLaVA-OneVision-72B~\citep{li2024llava}
& 43.5 & 23.9 & 57.6 & 37.5 & 42.5 & 39.9 & 32.5 & 44.6 & 40.2 \\
LLaVA-NeXT-Video-72B~\citep{liu2024llavanext}
& 48.9 & 22.8 & 57.4 & 35.3 & 42.4 & 36.7 & 35.0 & 48.6 & 40.9 \\
Qwen2.5-VL-72B~\citep{bai2025qwen25vltechnicalreport}
& 25.1 & 29.3 & 54.5 & 38.8 & 38.2 & 37.0 & 34.0 & 28.9 & 37.0 \\
Qwen3-VL-8B~\citep{bai2025qwen3}
& 67.5 & 47.0 & \underline{76.3} & 61.9 & 58.0 & 50.9 & 35.0 & 66.3 & 57.9 \\
InternVL3-8B~\citep{zhu2025internvl3}
& 68.1 & 39.0 & 48.4 & 33.6 & 48.3 & 36.4 & 27.3 & 35.4 & 42.1 \\
InternVL3-78B~\citep{zhu2025internvl3}
& 71.2 & \underline{53.7} & 44.4 & 39.5 & 55.9 & 39.5 & 28.9 & 54.5 & 48.4 \\

\tightmidrule
\rowcolor{GroupBlue}
\multicolumn{10}{c}{{\bfseries\itshape Spatial-Centric Models}} \\
Spatial-MLLM-4B~\citep{wu2026spatial}
& 65.3 & 34.8 & 63.1 & 45.1 & 41.3 & 46.2 & 33.5 & 46.3 & 48.4 \\
VG-LLM-4B~\citep{zheng2026learning}
& 66.0 & 37.8 & 55.2 & 59.2 & 44.6 & 45.6 & 33.5 & 36.4 & 47.3 \\
VG-LLM-8B~\citep{zheng2026learning}
& 67.9 & 37.7 & 58.6 & 62.0 & 46.6 & 40.7 & 32.4 & 59.2 & 50.7 \\
VST-3B-SFT~\citep{yang2025visual}
& 69.3 & 45.4 & 71.8 & 62.4 & 59.0 & 46.0 & 38.7 & 70.2 & 57.9 \\
VST-7B-SFT~\citep{yang2025visual}
& \underline{72.0} & 44.4 & 74.3 & \underline{68.3} & 59.7 & 55.8 & 44.9 & 65.2 & 60.6 \\
Cambrian-S-3B~\citep{yang2025cambrian}
& 70.7 & 40.6 & 68.0 & 46.3 & 64.8 & 61.9 & 27.3 & \textbf{78.8} & 57.3 \\
Cambrian-S-7B~\citep{yang2025cambrian}
& 68.2 & 45.8 & 72.5 & 67.6 & \underline{66.8} & 69.6 & 39.2 & \underline{73.8} & \underline{62.9} \\
VLM-3R-7B~\citep{fan2025vlm}
& 70.2 & 49.4 & 69.2 & 67.1 & 65.4 & \underline{80.5} & 45.4 & 40.1 & 60.9 \\
SpatialLadder-3B~\citep{li2025spatialladder}
& 62.1 & 35.3 & 61.9 & 41.4 & 45.6 & 46.4 & 27.3 & 38.5 & 44.8 \\

\tightmidrule
LongSpace-9B (Ours)
& \textbf{73.8} & \textbf{57.6} & \textbf{77.4} & \textbf{74.4} & \textbf{72.4} & \textbf{84.3} & 47.4 & \textbf{78.8} & \textbf{70.8} \\

\bottomrule
\vspace{-1.6em}
\end{tabular*}
\end{table*}

Table~\ref{tab:longspacebench_results} compares models on LongSpace-Bench. LongSpace-9B achieves the highest overall score of 49.2, outperforming the strongest open-source baseline, Qwen3-VL-32B~\citep{bai2025qwen3}, and the strongest proprietary baseline, Gemini-3-Pro~\citep{comanici2025gemini}, by 2.7 and 3.9 points, respectively. The gap is larger over spatial-centric models, with gains of 8.7 points over Cambrian-S-7B~\citep{yang2025cambrian} and 9.0 points over VLM-3R-7B~\citep{fan2025vlm}. At the task level, LongSpace-9B performs best on Appearance Order, State Change, and Route Recall, with scores of 52.5, 44.9, and 51.8, and ties for the best score on Relative Orientation. The gains, however, are not uniform across categories. Gemini-3-Pro remains stronger on Scene Classification, Scene Consistency, Relative Distance, Egocentric Reasoning, and Route Planning. These results suggest that explicit long-horizon memory is most useful when the task requires retaining and retrieving evidence from distant video segments, whereas some scene-level recognition and high-level planning questions still benefit from stronger proprietary models.
\begin{table}[H]
\centering
\caption{Effect of the number of geometry-injection layers on benchmark performance. \protect\raisebox{0.15ex}{\protect\colorbox{GroupBlue}{\protect\phantom{\protect\rule{0.75em}{0.32em}}}} highlights the best result for each metric.}
\vspace{-0.80em}
\label{tab:ablation_layers}
\renewcommand{\arraystretch}{0.95}
\footnotesize

\resizebox{\columnwidth}{!}{%
\begin{tabular}{@{\hspace{2pt}}ccccc@{\hspace{2pt}}}
\toprule
\textbf{\#Layers} &
\textbf{VSI-Bench} &
\textbf{CV-Bench} &
\textbf{SPAR-Bench} &
\textbf{Avg.} \\
\midrule
6  
& 63.6 
& 86.5 
& 65.5 
& 71.9 \\

8  
& 65.0 
& \cellcolor{GroupBlue}86.8 
& 65.3 
& \cellcolor{GroupBlue}72.4 \\

12 
& \cellcolor{GroupBlue}65.0 
& 86.4 
& 65.4 
& 72.3 \\

24 
& 61.3 
& 85.2 
& \cellcolor{GroupBlue}67.3 
& 71.3 \\

36 
& 64.9 
& 86.1 
& 65.3 
& 72.1 \\
\bottomrule
\end{tabular}%
}
\vspace{-1.45em}
\end{table}

\subsection{Ablation Studies}
\noindent\textbf{Moderate Geometry Injection.} Table~\ref{tab:ablation_layers} examines how many decoder layers receive geometry injection. All variants are trained on the same 324K training subset. The best average score is achieved with 8 layers, reaching 72.4, while 12 layers gives a comparable score of 72.3. Adding geometry to more layers does not improve performance monotonically. Although using 24 layers yields the highest SPAR-Bench~\citep{zhang2026flatland} score of 67.3, its average score drops to 71.3 because the scores on VSI-Bench and CV-Bench~\citep{tong2024cambrian} decrease to 61.3 and 85.2, respectively. This trend suggests that geometry injection is most effective when applied to a moderate number of early decoder layers.

\begin{table}[htbp]
\centering
\caption{Component-wise ablation of layer-aware memory organization, capacity allocation, and hierarchical memory roles.}
\label{tab:ablation_layer_aware}
\label{tab:ablation_memory_budget_roles}
\vspace{-0.6em}
\setlength{\tabcolsep}{2.0pt}
\renewcommand{\arraystretch}{0.94}
\scriptsize

\newcommand{\ablationgrouprow}[1]{%
\noalign{\vskip -1.00ex}%
\rowcolor{GroupBlue}[0pt][0pt]
\multicolumn{5}{@{}l@{}}{\rule{0pt}{2.05ex}\hspace{0.25em}\textit{#1}}\\[0.20ex]
}

\begin{tabular*}{\columnwidth}{@{\extracolsep{\fill}}lcccc@{}}
\toprule
\textbf{Configuration} &
\begin{tabular}[c]{@{}c@{}}
\textbf{Scene}\\
\textbf{Perception}
\end{tabular} &
\begin{tabular}[c]{@{}c@{}}
\textbf{Spatial}\\
\textbf{Relation}
\end{tabular} &
\begin{tabular}[c]{@{}c@{}}
\textbf{Spatial}\\
\textbf{Memory}
\end{tabular} &
\textbf{Overall} \\
\midrule
\ablationgrouprow{Layer organization}
Layer-agnostic
& 38.5 & 42.7 & 43.1 & 41.8 \\
Layer-aware
& 50.9 & 46.0 & 49.5 & 49.2 \\
\midrule
\ablationgrouprow{Memory budget}
Read cap. $\times 0.5$
& 36.7 & 35.7 & 35.6 & 35.9 \\
Read cap. $\times 2$
& 37.0 & 35.4 & 35.2 & 35.8 \\
Bank cap. $\times 0.5$
& 37.0 & 35.4 & 35.2 & 35.8 \\
\midrule
\ablationgrouprow{Hierarchical role design}
w/o working role
& 36.0 & 36.5 & 35.4 & 35.8 \\
w/o long role
& 34.5 & 35.9 & 34.6 & 34.8 \\
\bottomrule
\vspace{-1.5em}
\end{tabular*}
\end{table}

\begin{figure}[htbp]
    \centering
    \vspace{-0.55em}
    \includegraphics[width=\columnwidth,page=1]{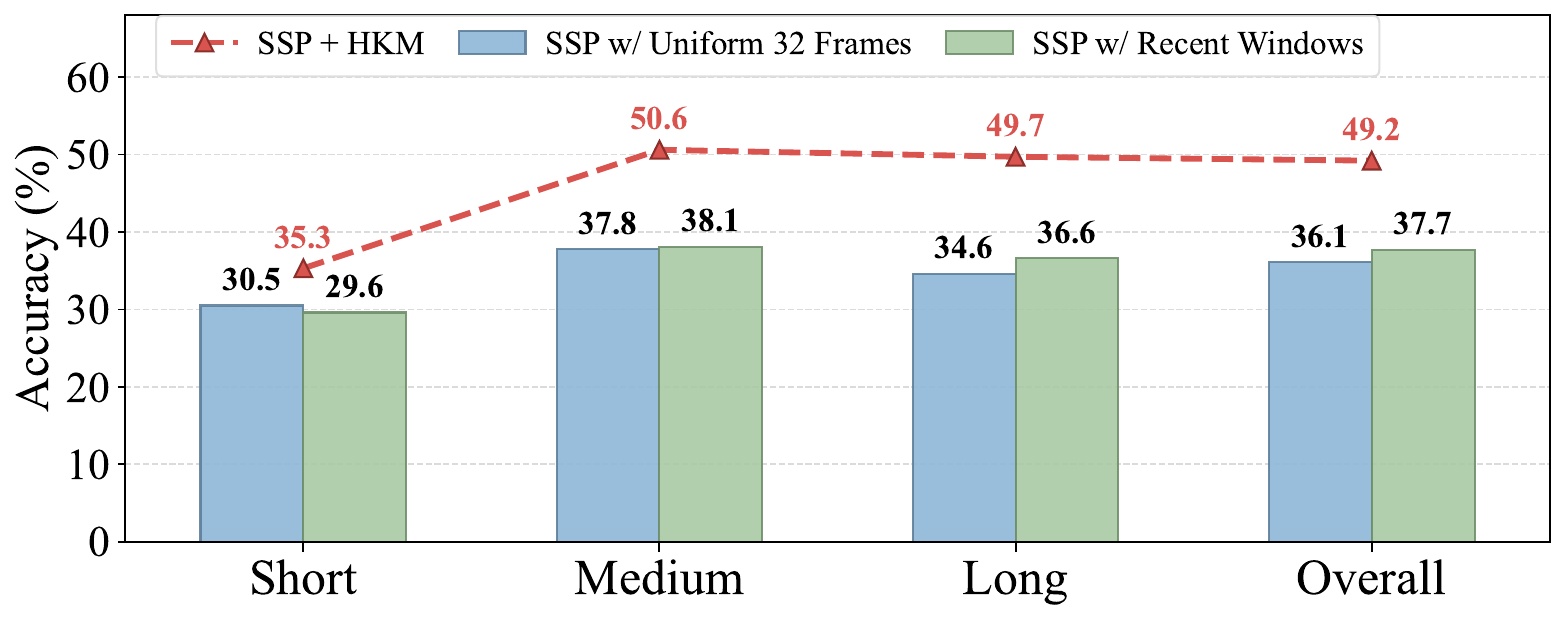}
    \vspace{-1.90em}
    \caption{Comparison of different inference settings on LongSpace-Bench across video length levels.}
    \vspace{-1.35em}
    \label{fig:longspace_ablation_bar_line}
\end{figure}

\noindent\textbf{Long-Horizon Memory Organization.}
Figure~\ref{fig:longspace_ablation_bar_line} and Table~\ref{tab:ablation_layer_aware} analyze long-memory inference and layer-aware organization on LongSpace-Bench. Uniform sampling with 32 frames scores 36.1, while recent-window inference gives a modest gain to 37.7. Long-memory inference reaches 49.2, outperforming the two baselines by 13.1 and 11.5 points. The gain grows with video length, improving short, medium, and long videos by 4.8, 12.8, and 15.1 points over uniform sampling. This suggests that preserving cross-chunk evidence matters more when relevant observations are separated by longer temporal intervals. We further examine whether these gains come from memory organization or memory size. The layer-aware design improves the overall score from 41.8 to 49.2, with the largest gain on scene perception. In comparison, changing the read or memory-bank capacity around the default setting keeps the score near 35.8--35.9. Removing hierarchical roles degrades performance, especially without the long-term role, reducing the score to 34.8. These results show that effective long-video memory relies more on layer- and role-specific organization than on memory capacity alone.

\section{Conclusion}
We introduce LongSpace-Bench and LongSpace to study spatial reasoning and memory in long videos. LongSpace-Bench is a long-horizon spatial memory benchmark built from real-world room-tour videos, covering scene perception, spatial relations, and spatial memory. LongSpace combines spatial structure perception with hierarchical KV memory to preserve and retrieve question-relevant spatial evidence across video segments. Together, they provide a foundation for evaluating and improving spatial memory in video MLLMs.

\section*{Limitations}
LongSpace-Bench is built primarily from indoor room-tour videos, which contain rich object layouts, room transitions, and long-range spatial dependencies but do not fully cover complex outdoor environments. And the proposed tasks focus on observation-based spatial memory, requiring the model to remember objects, regions, routes, and temporal order from passively observed videos, while active exploration, interactive navigation, and object manipulation remain outside the current scope.

\bibliography{custom}

\clearpage
\appendix
\flushbottom
\setlength{\textfloatsep}{6pt plus 1pt minus 2pt}
\setlength{\floatsep}{6pt plus 1pt minus 2pt}
\setlength{\intextsep}{6pt plus 1pt minus 2pt}

\section{Appendix}
\label{sec:longspace_appendix}

\subsection{LongSpace-Bench Data}
\label{sec:appendix_bench_construction}

Table~\ref{tab:longspace_data_card} summarizes the core data properties of LongSpace-Bench. The benchmark is built from Youtube real indoor room-tour videos and keeps clips that contain layout, object-position, viewpoint-change, and route evidence. Each question-answer sample is grounded in video evidence and is assigned to one of three ability groups: scene perception, spatial relationship, and spatial memory.

\begin{table}[h]
\centering
\caption{Data card of LongSpace-Bench.}
\label{tab:longspace_data_card}
\footnotesize
\resizebox{\columnwidth}{!}{%
\begin{tabular}{@{}ll@{}}
\toprule
\textbf{Item} & \textbf{Description} \\
\midrule
Video source & YouTube real-world indoor room-tour videos \\
Number of videos & 445 \\
Total duration & Approximately 159 hours \\
Number of QA pairs & 4,073 \\
Task categories & Scene perception, Spatial relationship, Spatial memory \\
Number of subtasks & 10 \\
Length subsets & Short: 280 QA; Medium: 2,102 QA; Long: 1,691 QA \\
Answer format & Numeric answers for counting; multiple choice otherwise \\
Evaluation metric & MRA for counting; accuracy for multiple choice \\
\bottomrule
\end{tabular}%
}
\end{table}

\begin{figure*}[htbp]
    \centering
    \includegraphics[width=\textwidth,page=1]{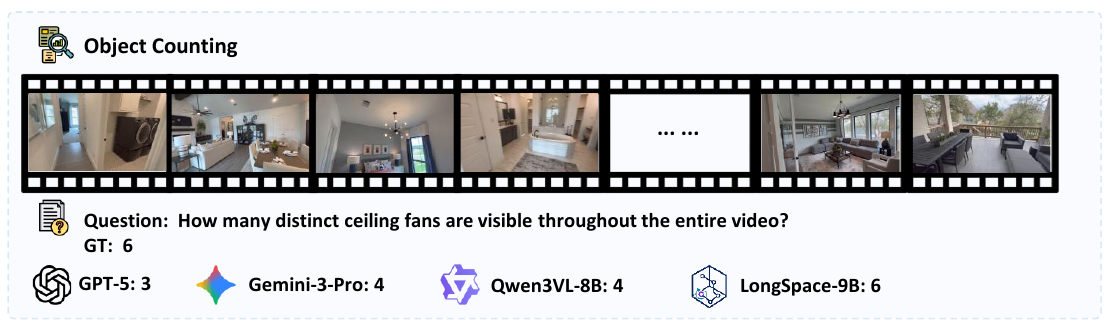}
    \caption{Object Counting QA Example in LongSpace-Bench}
    \label{fig:bench_object_counting}
\end{figure*}

\begin{figure*}[htbp]
    \centering
    \includegraphics[width=\textwidth,page=1]{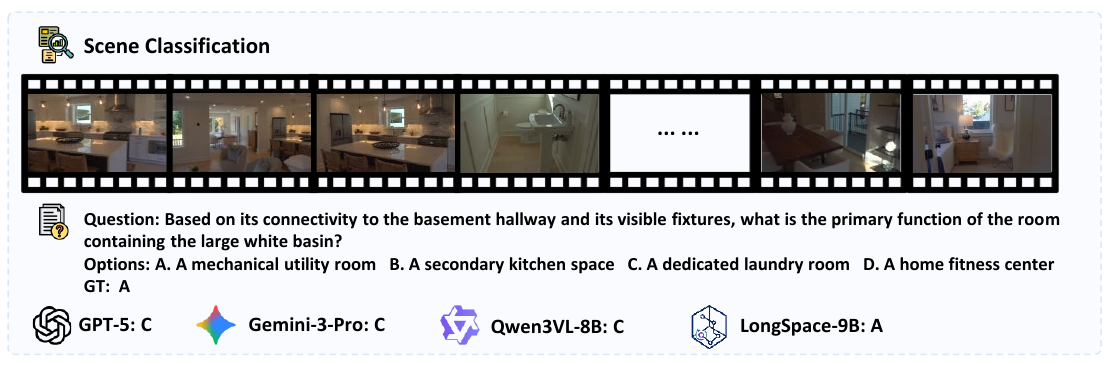}
    \caption{Scene Classification QA Example in LongSpace-Bench}
    \label{fig:bench_scene_classification}
\end{figure*}

\begin{figure*}[htbp]
    \centering
    \includegraphics[width=\textwidth,page=1]{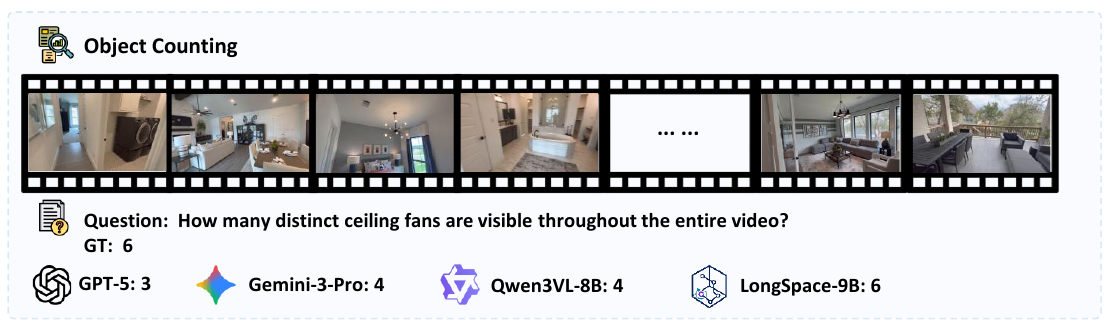}
    \caption{Scene Consistency QA Example in LongSpace-Bench}
    \label{fig:bench_scene_consistency}
\end{figure*}

\begin{figure*}[htbp]
    \centering
    \includegraphics[width=\textwidth,page=1]{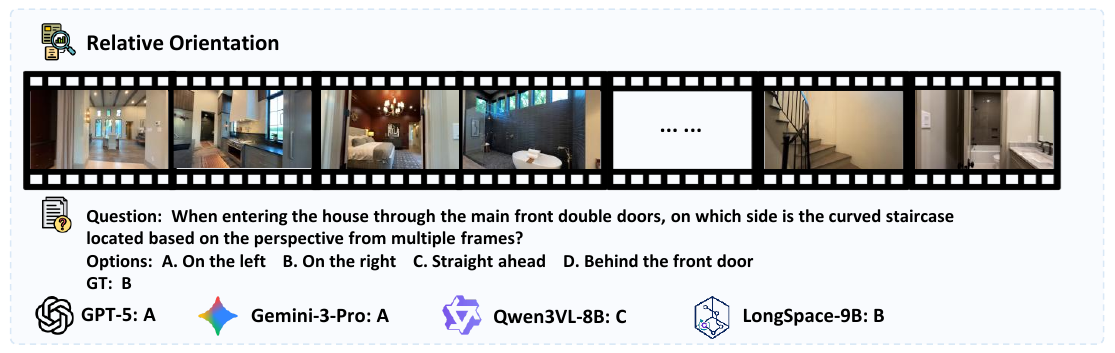}
    \caption{Relative Orientation QA Example in LongSpace-Bench}
    \label{fig:bench_relative_orientation}
\end{figure*}

\begin{figure*}[htbp]
    \centering
    \includegraphics[width=\textwidth,page=1]{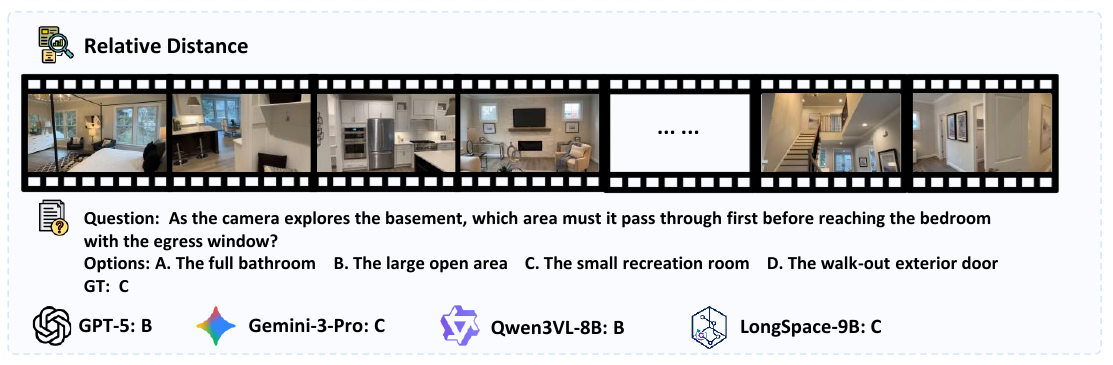}
    \caption{Relative Distance QA Example in LongSpace-Bench}
    \label{fig:bench_relative_distance}
\end{figure*}

\begin{figure*}[htbp]
    \centering
    \includegraphics[width=\textwidth,page=1]{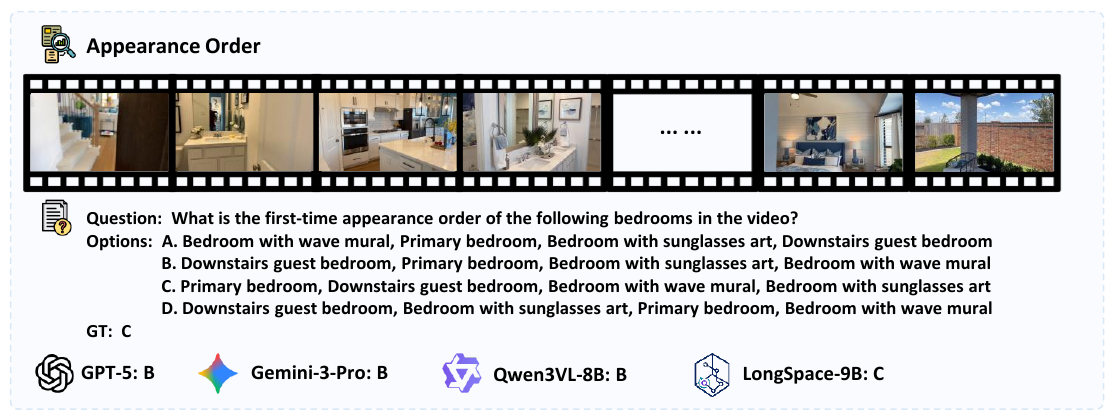}
    \caption{Appearance Order QA Example in LongSpace-Bench}
    \label{fig:bench_appearance_order}
\end{figure*}

\begin{figure*}[htbp]
    \centering
    \includegraphics[width=\textwidth,page=1]{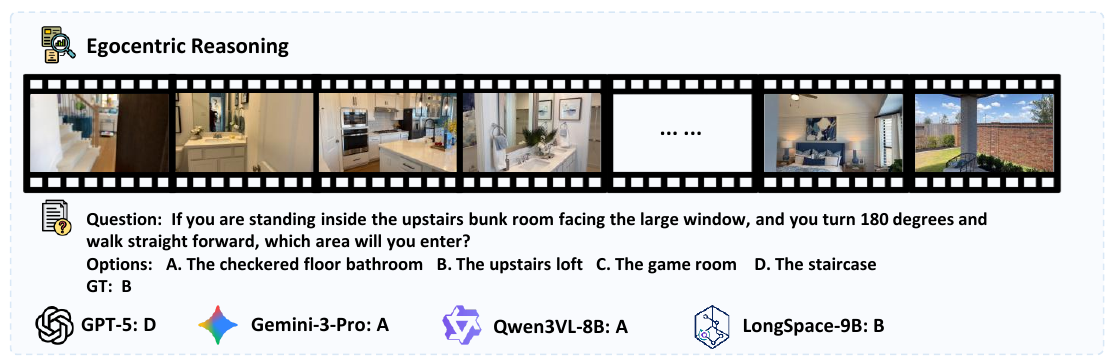}
    \caption{Egocentric Reasoning QA Example in LongSpace-Bench}
    \label{fig:bench_egocentric_reasoning}
\end{figure*}

\begin{figure*}[htbp]
    \centering
    \includegraphics[width=\textwidth,page=1]{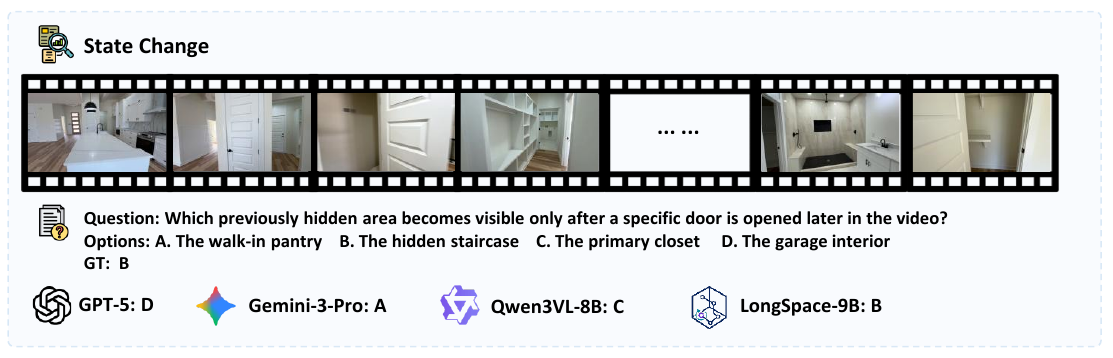}
    \caption{State Change QA Example in LongSpace-Bench}
    \label{fig:bench_stage_change}
\end{figure*}

\begin{figure*}[htbp]
    \centering
    \includegraphics[width=\textwidth,page=1]{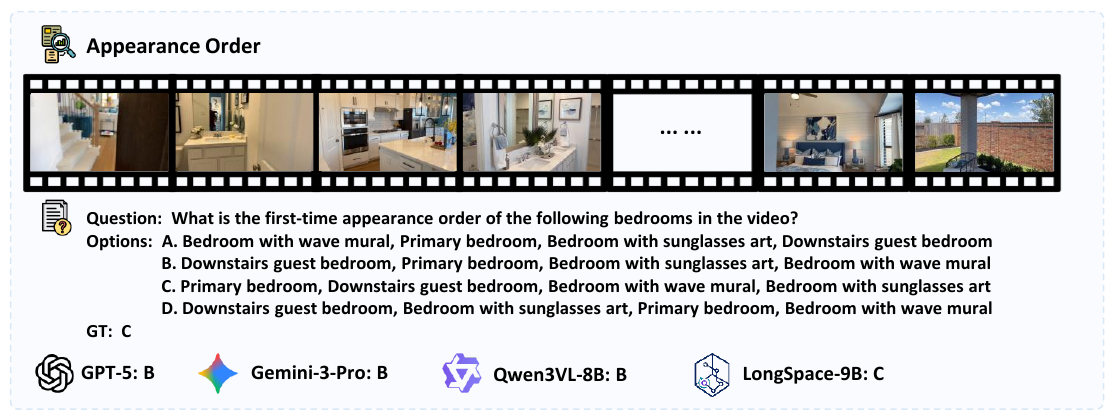}
    \caption{Route Planning QA Example in LongSpace-Bench}
    \label{fig:bench_route_planning}
\end{figure*}

\begin{figure*}[htbp]
    \centering
    \includegraphics[width=\textwidth,page=1]{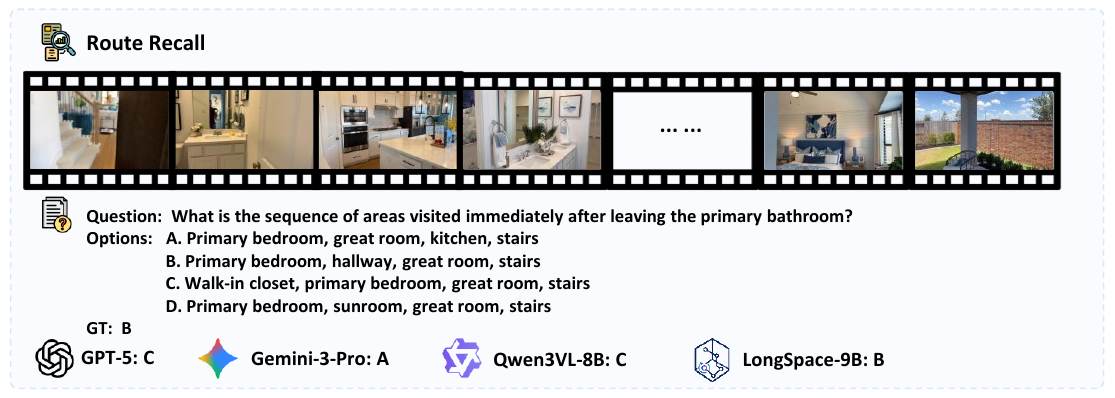}
    \caption{Route Recall QA Example in LongSpace-Bench}
    \label{fig:bench_route_recall}
\end{figure*}

Table~\ref{tab:longspace_task_stats} further reports the sample count, answer format, and evidence scope for each task type. Object Counting uses numerical answers, while the remaining tasks mainly use multiple-choice answers. This mixed format separates fine-grained quantitative perception from categorical spatial reasoning and memory retrieval. The QA examples for each LongSpace-Bench subtask are presented in Figures~\ref{fig:bench_object_counting}--\ref{fig:bench_route_recall}.

\begin{table}[H]
\centering
\caption{Task taxonomy and sample statistics of LongSpace-Bench. NA denotes numerical answer and MC denotes multiple choice.}
\label{tab:longspace_task_stats}
\scriptsize
\resizebox{\columnwidth}{!}{%
\begin{tabular}{@{}llcl@{}}
\toprule
\textbf{Ability} & \textbf{Task} & \textbf{QA} & \textbf{Evidence focus} \\
\midrule
\multirow{3}{*}{Scene perception}
& Object Counting & 500 / NA & Countable objects or regions \\
& Scene Classification & 367 / MC & Room or scene category \\
& Scene Consistency & 321 / MC & Stable scene identity and layout \\
\midrule
\multirow{2}{*}{Spatial relationship}
& Relative Distance & 387 / MC & Object or region distance \\
& Relative Orientation & 516 / MC & Left/right/front/back relation \\
\midrule
\multirow{5}{*}{Spatial memory}
& Appearance Order & 514 / MC & Temporal order of observations \\
& State Change & 285 / MC & Updated scene or object state \\
& Egocentric Reasoning & 421 / MC & Viewer-centered direction and location \\
& Route Planning & 293 / MC & Path-level spatial decision \\
& Route Recall & 469 / MC & Remembered navigation trajectory \\
\bottomrule
\end{tabular}%
}
\end{table}

\subsection{Annotation and Filtering Protocol}
\label{sec:appendix_annotation_workflow}

\begin{figure*}[!t]
    \centering
    \includegraphics[width=\textwidth,page=1]{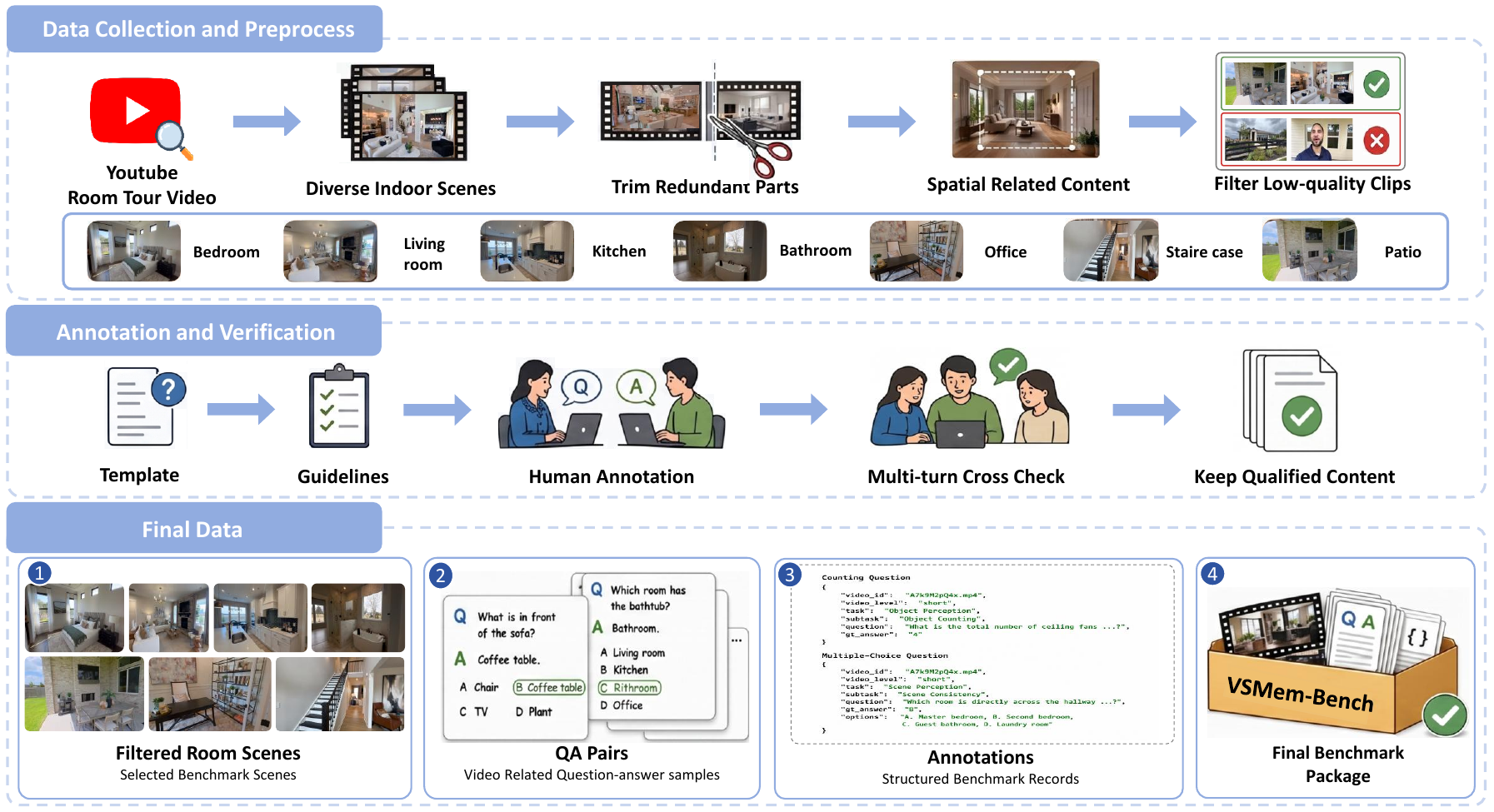}
    \caption{Construction pipeline of LongSpace-Bench. The pipeline collects room-tour videos, removes clips unrelated to spatial reasoning and memory, and produces verified question-answer pairs through taxonomy-guided annotation and manual review.}
    \label{fig:bench_pipeline}
\end{figure*}

Figure~\ref{fig:bench_pipeline} shows the construction pipeline of LongSpace-Bench. We start from raw YouTube videos and clip them into relevant segments by annotating their start and end timestamps and removing portions unrelated to spatial reasoning. The retained clips are arranged as temporally ordered observation sequences, so that questions can refer to evidence across different rooms, viewpoints, and time steps.

We provide question templates and descriptions of common indoor objects and regions. Annotators write questions, candidate answers, and ground-truth answers according to the corresponding task type. During review, we check whether the supporting evidence can be located in the video, whether each answer is unique, and whether the referring expressions are clear. We remove samples if the answer cannot be uniquely determined from the video, the question requires external commonsense knowledge, the target object is heavily occluded, or the options conflict ambiguously with the video evidence.

\subsection{Answer Format and Quality Control}
\label{sec:appendix_answer_format}

Object Counting uses numerical answers and is evaluated with mean relative accuracy (MRA), which limits the impact of small counting deviations while still penalizing large errors. The remaining tasks mainly use multiple-choice answers and are evaluated by accuracy. These tasks cover relation judgment, state recall, route reasoning, and temporal-order reasoning.

Quality control aims to reduce shortcuts and ambiguity. For numerical questions, annotators check whether the target to be counted is clearly defined and visible in the video. For multiple-choice questions, distractors are written to be plausible in the room context but inconsistent with the video evidence. This design encourages models to retrieve the relevant spatial observation instead of relying on language priors or local static cues.

\subsection{Evaluated Models and Input Protocol}
\label{sec:appendix_experimental_details}

Table~\ref{tab:eval_protocol} summarizes the input protocol used for each evaluated model. We follow the practical input constraints of each model interface and provide temporally distributed evidence whenever possible. For proprietary models, GPT-5 is evaluated with 50 uniformly sampled frames, while Gemini-3-Pro is evaluated with video input sampled at 1 fps. For open-source video MLLMs, the Qwen series is evaluated with 512 uniformly sampled frames over the full video. InternVL3.5, LLaVA-OneVision-1.5, LLaVA-NeXT-Video, LongVA, and LongVILA are evaluated with 64 uniformly sampled frames. For spatial-centric MLLMs, we follow the default configurations of each model: Spatial-MLLM-4B uses 16 frames, while VG-LLM, VST, VLM-3R, SpatialLadder, and Cambrian-S use 32 frames. LongSpace uses chunked memory inference, where videos are processed at 1 fps and divided into 32-frame chunks with a 4-frame overlap.

For multiple questions from the same video, LongSpace constructs the video memory once and reuses it across different question prompts. This avoids repeated encoding of the full long video and keeps the evaluation protocol consistent with the long-memory inference setting.

\begin{table}[t]
\centering
\caption{Evaluated models and input protocols.}
\label{tab:eval_protocol}
\scriptsize
\begin{tabularx}{\columnwidth}{@{}>{\raggedright\arraybackslash}p{3.6cm}X@{}}
\toprule
\textbf{Model} & \textbf{Input protocol} \\
\midrule
GPT-5 & 50 uniformly sampled frames \\
Gemini-3-Pro & 1 fps video input \\
\midrule
Qwen series & 512 uniformly sampled frames \\
InternVL3.5 & 64 uniformly sampled frames \\
LLaVA-OneVision-1.5 & 64 uniformly sampled frames \\
LLaVA-NeXT-Video & 64 uniformly sampled frames \\
LongVA & 64 uniformly sampled frames \\
LongVILA & 64 uniformly sampled frames \\
\midrule
Spatial-MLLM-4B~\citep{wu2026spatial} & 16 frames under the default setting \\
VG-LLM-4B~\citep{zheng2026learning} & 32 frames under the default setting \\
VG-LLM-8B~\citep{zheng2026learning} & 32 frames under the default setting \\
VST-3B-SFT~\citep{yang2025visual} & 32 frames under the default setting \\
VST-7B-SFT~\citep{yang2025visual} & 32 frames under the default setting \\
VLM-3R-7B~\citep{fan2025vlm} & 32 frames under the default setting \\
SpatialLadder-3B~\citep{li2025spatialladder} & 32 frames under the default setting \\
Cambrian-S-3B~\citep{yang2025cambrian} & 32 frames under the default setting \\
Cambrian-S-7B~\citep{yang2025cambrian} & 32 frames under the default setting \\
\midrule
\raisebox{-1.8ex}{LongSpace} &
1 fps chunked memory inference; \newline
32 frames per chunk; \newline
4-frame overlap \\
\bottomrule
\end{tabularx}
\end{table}

\subsection{Training Data Sources}
\label{sec:appendix_training_data}

Our spatial instruction tuning data are drawn from three complementary sources: VSI-590K, VLM-3R, and a sampled subset of SPAR-7M, denoted as SPAR-234K. This data mixture follows the training data setting of VG-LLM, with VSI-590K further included to strengthen video-level spatial supervision. This mixture provides broad spatial supervision, including object-level perception, geometric relation reasoning, and video-level spatial understanding. Overall, it contains approximately 1.18M spatial instruction samples.

\textbf{VSI-590K.} VSI-590K is a large-scale spatial instruction dataset designed for visual-spatial understanding. We use this dataset to provide video-centric supervision for indoor spatial reasoning, including object perception, distance estimation, relative direction, route reasoning, and appearance-order understanding. These samples help the model learn spatial concepts that are directly tied to observations over video frames.

\textbf{VLM-3R.} VLM-3R introduces instruction data for 3D-aware visual reasoning from monocular video. Its supervision emphasizes spatial context, scene structure, camera motion, and geometry-language alignment. We use the VLM-3R instruction data to strengthen the ability of the model to connect visual observations with implicit 3D structure, which is important for reasoning about layouts and viewpoint changes.

\textbf{SPAR-234K.} SPAR-234K is a sampled subset of SPAR-7M used for spatial reasoning instruction tuning. SPAR-7M is built from scenes with 3D ground truth and covers diverse spatial tasks, ranging from basic perception to relation and layout reasoning. Following the subset setting used in VG-LLM, we sample approximately 234K examples to broaden the distribution of spatial relations while keeping the training mixture compact.

\begingroup
\SetAlCapNameFnt{\small}
\SetAlCapFnt{\small}
\SetKwInput{KwNotation}{Notation}

\begin{algorithm}[t]
\small
\SetAlgoNoLine
\DontPrintSemicolon
\caption{Long-Video Inference with Hierarchical KV Memory (HKM)}
\label{alg:kv_memory}

\KwIn{Video $\mathcal{V}$, final question $q$, segment size $C$, memory budgets $\{B_{\rho}\}$, read budgets $\{R_{\rho}\}$}
\KwOut{Answer $a$}
\KwNotation{$\mathcal{V}$: input video; $q$: final question; $C$: frames per segment; $T$: number of ordered segments; $a$: generated answer.}
\KwNotation{$l$: decoder-layer index; $\rho(l)$: role of layer $l$ (sensory, working, or long-memory).}
\KwNotation{$\mathcal{M}_{t,l}$: memory at layer $l$ after segment $t$; $\mathcal{A}_{t,l}$: entries selected from segment $t$.}
\KwNotation{$(\mathbf{K}_{t,l},\mathbf{V}_{t,l})$: KV states of segment $t$; $\mathbf{F}_{t,l}$: hidden features used for memory scoring.}
\KwNotation{$\mathbf{e}_q$: question embedding; $(\mathbf{K}^{\mathrm{mem}}_l,\mathbf{V}^{\mathrm{mem}}_l)$: retrieved KV states; $B_{\rho(l)}$/$R_{\rho(l)}$: role-specific memory/read budgets.}

Sample frames from $\mathcal{V}$ and split them into ordered segments $\{\mathcal{C}_t\}_{t=1}^{T}$ of size $C$\;
Initialize HKM states $\mathcal{M}_{0,l}=\emptyset$ for each decoder layer $l$\;

\For{$t=1$ \KwTo $T$}{
    Encode segment $\mathcal{C}_t$ with Spatial Structure Perception (SSP)\;

    \ForEach{decoder layer $l$}{
        Obtain current-segment states
        $(\mathbf{K}_{t,l},
        \mathbf{V}_{t,l},
        \mathbf{F}_{t,l})$\;

        Select position-preserving HKM entries:
        $\mathcal{A}_{t,l}
        =
        \operatorname{Select}_{\rho(l)}
        (\mathbf{K}_{t,l},
        \mathbf{V}_{t,l},
        \mathbf{F}_{t,l})$\;

        Update HKM:
        $\mathcal{M}_{t,l}
        =
        \operatorname{Compress}_{\rho(l)}
        (\mathcal{M}_{t-1,l}\cup\mathcal{A}_{t,l};
        B_{\rho(l)})$\;
    }
}

Compute query embedding $\mathbf{e}_q$ from the final question $q$\;

\ForEach{decoder layer $l$}{
    Retrieve HKM states:
    $(\mathbf{K}^{\mathrm{mem}}_{l},
    \mathbf{V}^{\mathrm{mem}}_{l})
    =
    \operatorname{Read}_{\rho(l)}
    (\mathcal{M}_{T,l},\mathbf{e}_q;R_{\rho(l)})$\;
}

Generate answer $a$ from $q$ using the retrieved HKM states as frozen memory prefixes\;

\Return{$a$}\;

\end{algorithm}

\endgroup

\subsection{Long-Video Inference with Hierarchical KV Memory}
\label{sec:appendix_inference_algorithm}

Algorithm~\ref{alg:kv_memory} provides the Hierarchical KV Memory (HKM) procedure used by LongSpace during inference. The model first samples and splits the video into temporally ordered segments, then encodes each segment with Spatial Structure Perception (SSP). At each decoder layer, HKM selects current KV states and hidden scoring features according to the layer role, and compresses them into the corresponding hierarchical memory.

\noindent\textbf{Notation.}
Let $\mathcal{V}$ denote the input video, $q$ the final question, $C$ the number of frames per segment, and $\{\mathcal{C}_t\}_{t=1}^{T}$ the resulting sequence of $T$ video segments. Following Section~\ref{sec:hkm}, $l$ indexes a decoder layer and $\rho(l)$ denotes its role (sensory, working, or long-memory). $B_{\rho(l)}$ specifies the memory budget for that role, while $R_{\rho(l)}$ denotes the number of entries allowed during retrieval. The generated answer is denoted by $a$.

For segment $t$ at layer $l$, $\mathbf{K}_{t,l}$ and $\mathbf{V}_{t,l}$ are the KV states produced by the current segment, and $\mathbf{F}_{t,l}$ is the hidden feature used for scoring, consistently with Eqs.~\ref{eq:memory_select}--\ref{eq:memory_update}. $\operatorname{Select}_{\rho(l)}$ forms the candidate set $\mathcal{A}_{t,l}$, preserving temporal position information, and $\operatorname{Compress}_{\rho(l)}$ merges it with $\mathcal{M}_{t-1,l}$ to obtain the updated memory $\mathcal{M}_{t,l}$ under budget $B_{\rho(l)}$. After all $T$ segments are written, $\mathbf{e}_q$ denotes the embedding of the final question, and $\operatorname{Read}_{\rho(l)}$ retrieves $(\mathbf{K}^{\mathrm{mem}}_l,\mathbf{V}^{\mathrm{mem}}_l)$ from $\mathcal{M}_{T,l}$ under budget $R_{\rho(l)}$. These retrieved KV states remain frozen during generation and are prepended to the autoregressive cache. Thus, HKM changes inference-time memory construction and retrieval without introducing a new attention operator or an additional training objective.

\end{document}